\documentclass[runningheads]{llncs}

\usepackage{booktabs}           
\usepackage{multirow}           
\usepackage{amsfonts}           
\usepackage{graphicx}           
\usepackage{duckuments}         

\usepackage[numbers,compress,sort]{natbib}

\title{Predicting Drug-Drug Interactions Using Knowledge Graphs}
\titlerunning{Predicting Drug-Drug Interactions Using Knowledge Graphs}

\author{Lizzy Farrugia\inst{1} \and
Lilian M. Azzopardi\inst{2} \and
Jeremy Debattista\inst{1} \and
Charlie Abela\inst{1}}
\authorrunning{Lizzy Farrugia et al.}
\institute{Department of Artificial Intelligence, Faculty of ICT, University of Malta\\
\email{lizzy.farrugia@um.edu.mt, jerdebattista@gmail.com, charlie.abela@um.edu.mt}\\
\and
{Department of Pharmacy, Faculty of Medicine and Surgery, University of Malta\\ 
\email{lilian.azzopardi@um.edu.mt}\\
}}

\begin{document}

\maketitle

\begin{abstract}
In the last decades, people have been consuming and combining more drugs than
before, increasing the number of Drug-Drug Interactions (DDIs). To predict
unknown DDIs, recently, studies started incorporating Knowledge Graphs (KGs)
since they are able to capture the relationships among entities providing better
drug representations than using a single drug property.
In this paper, we propose the medicX end-to-end framework that integrates
several drug features from public drug repositories into a KG and embeds the nodes in the
graph using various translation, factorisation and Neural Network (NN) based KG
Embedding (KGE) methods. Ultimately, we use a Machine Learning (ML) algorithm
that predicts unknown DDIs. Among the different translation and
factorisation-based KGE models, we found that the best performing combination
was the ComplEx embedding method with a Long Short-Term Memory (LSTM) network,
which obtained an $F_{1}$-score of 95.19\% on a dataset based on the DDIs found
in DrugBank version 5.1.8. This score is 5.61\% better than the state-of-the-art
model DeepDDI \cite{ryu}. Additionally, we also developed a graph auto-encoder
model that uses a Graph Neural Network (GNN), which achieved an $F_{1}$-score of
91.94\%. Consequently, GNNs have demonstrated a stronger ability to mine the
underlying semantics of the KG than the ComplEx model, and thus using higher
dimension embeddings within the GNN can lead to state-of-the-art performance.
\end{abstract}

\section{Introduction}
\label{intro}

Drug-Drug Interactions (DDIs) occur when two or more medications are co-administered simultaneously and cause an Adverse Drug Reaction (ADR). An estimated 44\% of men and 57\% of women older than 65 in the US take five or more medications, which is bound to worsen, given the population’s rapid ageing and the trend of increasing medication use \cite{woodruff2010preventing}. Moreover, the risk of an ADR increases by 7 to 10\% with each medication \cite{garber2019medication}.

Currently, drug developers rely on clinical trials to detect unknown DDIs,
whilst pharmacists and doctors depend on a textbook, such as the British
National Formulary (BNF), when they are unsure of a known DDI. Amran et al.
\cite{amran} reported that less than 50\% of ADRs are usually detected during
these trials since these are very labour-intensive, time-consuming, and
expensive, whilst the remaining ADRs are first-hand experienced by patients
during the post-marketing surveillance, known as PharmacoVigilance (PV).
Therefore, a framework that can predict potential DDIs, and serve as a reference
point to known and potentially dangerous DDIs is essential for healthcare
professionals to prescribe safe medications to patients, to provide better and
safer healthcare \cite{farrugia2020mining}.

Many researchers have focused on applying different Machine Learning (ML)
algorithms to predict unknown DDIs, including the state-of-the-art model DeepDDI
\cite{ryu}. However, this approach presents a limitation since drugs within
their deep network are represented using a single drug property, which is not
always available, and thus, their model cannot predict potential DDIs for such
instances. Furthermore, one property may not always be enough to represent a
drug. These drawbacks motivated other researchers, such as Celebi et al.
\cite{celebi}, to create better drug representations by building a Knowledge
Graph (KG) encompassing different drug relationships with other critical
drug-related concepts, such as diseases, targets, and genes. KGs bring the
ability to represent entities and relations with high reliability,
explainability, and reusability. These relations are then embedded into a single
feature using Knowledge Graph Embedding (KGE) methods. Various graph learning
methods exist, including Graph Neural Networks (GNNs) \cite{zitnik}, a
relatively new research area in Artificial Intelligence (AI), which has not been
extensively applied in the DDI domain.

\subsection{Aim \& Objectives}
\label{aim_objectives}

In this study, our aim is to build the medicX framework that accurately
predicts potential DDIs by leveraging KG and ML techniques. To address this aim,
we set two objectives.

\begin{itemize}
\item Objective 1 (O1): Generate a KG relating various drug-related concepts
such as drugs, diseases, side effects, and DDIs \cite{celebi, karim}. This first
entails the acquisition of different drug-related data from repositories such as
DrugBank \cite{drugbank} and SIDER \cite{sider}, since these are highly reputed
sources, and secondly, the integration and alignment of this data into a more
comprehensive KG. Although there are already other available biomedical KGs,
such as Bio2RDF \cite{bio2rdf}, the data has not been updated since 2014. To
evaluate this objective, we define a set of Competency Questions (CQs), a common
technique for ontology assessments \cite{bio2rdf}.
\item Objective 2 (O2): To predict new DDIs, we need to investigate, train and
fine-tune various KGE approaches using the KG mentioned in Objective $O1$ to
obtain high-quality drug feature vectors. Among these KGE approaches, we will
compare TransE, ComplEx and GNNs. In addition, an investigation will follow to
produce an accurate ML algorithm that relies on the obtained embeddings to
differentiate between an interacting and a non-interacting drug pair. Finally,
similar to other studies \cite{celebi, karim}, we plan to train and evaluate the
DDI predictor based on the positive DDIs found in the DrugBank version 5.1.8
dataset.
    
\end{itemize}

The rest of the paper is structured as follows: In Section 2, we present research related to KGs and unknown DDI prediction, and in Sections 3 and 4, we present the implementation details and evaluation methodology adopted to evaluate the solutions related to the objectives mentioned above, and finally, in Section 5, we provide some concluding remarks.

\section{Related Work}

In this section, we present research that adopted KGs to achieve their goals and different techniques that other works have implemented to predict DDIs. 

\subsection{Biomedical Knowledge Graphs}
With the ever-growing number of biomedical databases and the increasing popularity of semantic web technologies, there is a pressing need to develop systems that can integrate this data and offer an endpoint for users to query.

Himmelstein et al. \cite{hetionet} developed Hetionet, a KG developed for
Project Rephetio to systematically identify why drugs work and predict new
therapies for drugs. Furthermore, the Drug Repurposing Knowledge Graph
\cite{drkg2020} is a KG that builds upon Hetionet by integrating several
additional data resources and was initially developed as part of a project for
drug repurposing to target suitable treatments for COVID-19. Both graphs
comprise thousands of individual interconnections among various biomedical
concepts, such as diseases, side effects, and Anatomical Therapeutic Chemical
(ATC) Classification codes, extracted from several repositories, including
DrugBank and SIDER. Similarly, Zheng et al. \cite{pharmkb} developed PharmKG, a
multi-relational attributed biomedical KG, which contains thousands of gene,
compound, and disease nodes connected by a set of semantic relationships derived
from the abstracts of biomedical literature. Moreover, Abdelaziz et al.
\cite{tiresias} proposed Tiresias, a similarity-based system that encodes and
stores a KG in Resource Description Framework (RDF) format, then inputted into
Apache Spark for similarity calculation and model building for DDI prediction.

Bio2RDF \cite{bio2rdf} is a biological database that applies semantic web
technologies to publicly available databases to provide interlinked life science
data related to entities such as drugs, proteins, pathways and diseases. Their
method first entailed designing the ontology and converting public biomedical
datasets to RDF format documents. This process is known as rdfizing and was
achieved using Jakarta Server Pages (JSP), a technology that can create
dynamically generated web pages based on HTML and XML. Virtuoso Open Source, a
semantic web software, was then used to merge, query, and visualise the data.
Although the Bio2RDF KG stopped being maintained in 2014, bio2rdf-scripts,
written in PHP, are available for each dataset. The scripts convert the raw data
from the data sources to N-quad documents and have been adopted by several
researchers \cite{celebi, kgnn}.

\subsection{Predicting Unknown Drug-Drug Interactions}
\label{related_work:predicting_DDIs}
Traditionally, the discovery of DDIs relies on \textit{in vitro} and \textit{in
vivo} experiments and focuses on small sets of specific drug pairs in a
laboratory setting. However, with the emergence of available biomedical data and
since laboratory screenings of DDIs are very challenging and expensive, there is
growing interest in studying and predicting drug interactions using
computational methods.

At present, DDI prediction methods are mainly divided into similarity and
graph-based approaches. In similarity-based prediction, the underlying
assumption is that if \textit{Drug A} and \textit{Drug B} interact to produce a
specific biological effect, then drugs similar to \textit{Drug A} (or
\textit{Drug B}) are likely to interact with \textit{Drug B} (or \textit{Drug
A}) to produce the same effect \cite{vilar}.

Several studies \cite{vilar, indi} use similarity measures, which are functions
that take as input a particular drug property of \textit{Drug A} and
\textit{Drug B}, and calculate the similarity between them. The most popular
drug similarity measure is the Tanimoto Coefficient (TC) \cite{tanimoto}. The TC
calculates the similarity between the molecular fingerprints of the drug pair.
If the resultant TC is $0$ that means that the molecular structure of the drug
pair is \textit{maximally dissimilar} to one another, whilst if the TC is $1$,
then the drug pair is \textit{maximally similar}. However, creating a model that
solely relies on the TC does not yield high results \cite{vilar}.

In addition to the TC similarity measure, other researchers \cite{indi, ryu}
incorporated other similarity measures or ML approaches. For example, Ryu et al.
\cite{ryu} created a state-of-the-art framework called DeepDDI that can predict
DDIs and drug-food interactions. Besides the TC similarity measure to calculate
the similarity profile of drugs based on their SMILES notation, they also
employed a Principal Component Analysis (PCA) and a deep feed-forward NN to
reduce the feature space and predict the type of unknown drug-drug and drug-food
interactions.

A drawback of relying on a few features is that they may not always be available
and can be challenging and costly to obtain. A popular approach adopted by
multiple studies \cite{vilar, ryu} is to remove those biological entities
without features via pre-processing. However, this usually results in a
small-scale pruned dataset and thus is not pragmatic and useful in reality.
Furthermore, Vilar et al. \cite{vilar} observed that relying on a handful number
of features may not be precise enough to represent or characterise DDIs, and may
fail to help build a robust and accurate DDI model.

For these reasons, recent studies \cite{karim, tiresias} focus has turned
towards representing drug knowledge by leveraging a KG and then applying KGE
models to derive drug features by mapping nodes to a $d$-dimensional embedding
space so that similar nodes in the graph are embedded close to each other. These
models are typically categorised into translation, factorisation and NN-based
models.

Translation-based models, such as TransE \cite{TRANSE} and TransR \cite{TRANSR},
assume that after applying a relational translation when we add the embedding of
the head to the embedding of the relation, the result should be the embedding of
the tail entity. On the other hand, factorisation-based models, such as RESCAL
\cite{rescal}, DistMult \cite{distmult} and ComplEx \cite{complex}, capture
nodes and relations as multidimensional tensors, which are then factorised into
low-dimensional vectors. GNNs \cite{zitnik} are an active, new field of research
that use deep learning methods to perform inference on data described by graphs.
GNNs can be regarded as an embedding methodology that distils high-dimensional
information about each node’s neighbourhood into a dense vector embedding.

Abdelaziz et al. \cite{tiresias} compared different features, including word and
graph embedding-based features, which were calibrated and fed into a Logistic
Regression (LR) algorithm to predict potential DDIs. They concluded that HoIE
\cite{hole}, a factorisation-based KGE method, produced the most powerful
feature.

Furthermore, Celebi et al. \cite{celebi} and Karim et al. \cite{karim} evaluated
different translation and factorisation-based KGEs. From their experiments,
Celebi et al. \cite{celebi} concluded that RDF2Vec \cite{rdf2vec} performed the
best, whilst Karim et al. \cite{karim} concluded that translation-based models
have low expressive power since they do not capture semantic information, unlike
the ComplEx \cite{complex} KGE model, which obtained the best results in their
evaluation. Moreover, Celebi et al. \cite{celebi} demonstrated that Random
Forests (RFs) outperformed LR and Naive Bayes (NB) classifiers. Karim et al.
\cite{karim} confirmed Celebi et al.’s \cite{celebi} remarks but concluded that
a more complex ML algorithm, such as their Convolutional-LSTM model, can obtain
better results, achieving an $F_{1}$-score improvement of 7\% over the RF
classifier.

Recently, there has been an increasing interest in applying GNNs for DDI
prediction \cite{feng, zitnik}. Ji et al. \cite{ji2021survey} demonstrated that
such models have the generality to consider the type of entity and relation,
path information and underlying structure information and thus can resolve the
limitations of translation and factorisation-based models in representing all
the features of a KG.

For example, Feng et al. \cite{feng} developed Deep Predictor for DDIs (DPDDI),
which incorporates a Graph Convolutional Network (GCN) model that learns
low-dimensional feature representations of drugs by capturing the topological
relationship of drugs in a DDI network. A predictor then concatenates the latent
feature vectors of any two drugs and trains a 5-layer Deep Neural Network (DNN)
to predict potential DDIs.

Zitnik et al. \cite{zitnik} presented Decagon, a graph auto-encoder approach to
predict polypharmacy side effects. In their methodology, the authors collected
protein-protein, drug-protein and drug-drug interaction and side effect data.
Each DDI is labelled by a different edge type, which signifies the kind of side
effect, and hence, formulated the problem of predicting polypharmacy side
effects as solving a multi-relational link prediction task. They then developed
a multi-layer Relational-GCN (R-GCN), an extension of GCNs that can
differentiate between different types of relations in a KG to produce higher
quality embeddings. The R-GCN module acts as an encoder and operates on the
graph to produce embeddings for the nodes. A decoder in the form of a tensor
factorisation model then predicts a candidate edge, which is ultimately followed
by a sigmoid function to compute the probability that a given edge is one of the
given side effects.

\section{The medicX Approach}
In the previous section, we discussed research related to our objectives. In
this section, we discuss the implementation of the two main components of our
system, developed using the Python language.

\subsection{Building the Knowledge Graph}
We decided to create a KG using a multi-step approach that integrates drug data
from different bioinformatics sources to create a more comprehensive KG than the
existing ones that can ultimately infer new knowledge and generate high-quality drug features.

Similar to Celebi et al. \cite{celebi} and Karim et al. \cite{karim}, we
converted data from DrugBank \cite{drugbank}, KEGG \cite{kegg}, PharmGKB
\cite{pharmgkb} and SIDER \cite{sider} datasets to RDF, providing a more fluid
and effective model for integrating and querying the data using the
bio2rdf-scripts. The advantage of using this approach instead of using the
actual Bio2RDF KG or any existing KG is that this method helped us create a KG
that contains the most recent and up-to-date data.

Initially, we investigated the overlap among different reliable DDI
repositories, including DrugBank, KEGG, Drugs.com, Liverpool Covid-19
Interactions, TWOSIDES \cite{tatonetti2012data} and PharmGKB, and observed that
the overlap is minimal, highlighting the importance of merging DDIs from
multiple sources \cite{ayvaz}. Therefore, we decided to employ several data
mining tools, such as asynchronous requests, BeautifulSoup
\cite{richardson2007beautiful} and Selenium to scrape DDIs from these portals.
As various drug databases represent drugs using identifiers specific to them, we
mapped the drugs across different repositories using numerous techniques to
create a holistic list of unique DDIs. As a result, we managed to accumulate
2,477,864 unique DDIs.

Indications, also known as associated conditions, and ATC codes are frequently
used as features to represent the drugs in similarity-based approaches
\cite{indi}. For this reason, we scraped associated conditions from the DrugBank
portal, as these are not available in their XML dataset, and downloaded the ATC
Classification System \cite{skrbo2004classification} ontology, available from
BioPortal.

We then designed our ontology, which we refer to as the medicX ontology, to
encompass the extracted associated conditions and the DDIs, using the Web
Ontology Language (OWL) and RDF Schema (RDFS) ontology schemas through the
RDFLib \cite{rdflib} library. Finally, to enable interoperability in our KG, we
aligned the medicX ontology with the Bio2RDF and ATC Classification System
ontologies by determining correspondences among semantically related entities,
using relations such as \textit{owl:equivalentClass}, \textit{owl:sameAs} and
\textit{rdfs:subClassOf}. Then, we created a rdfizer that converts the mined
associated conditions and DDIs to RDF notation.

After performing the steps mentioned above, we imported the resulting processed
RDF documents into a triple store. In our study, we opted to use Ontotext’s
GraphDB\footnote{https://graphdb.ontotext.com/}, which can smoothly integrate
heterogeneous data from multiple sources and store hundreds of billions of facts
about any conceivable concept, unlike other research work that used alternative
graph databases such as the Blazegraph\footnote{https://blazegraph.com/}
Database, used by Karim et al. \cite{karim}, and Virtuoso Open
Source\footnote{https://virtuoso.openlinksw.com/}, which hosts the Bio2RDF KG.

\subsection{Predicting Unknown Drug-Drug Interactions}
In order to be able to predict DDIs, once the data is in the form of a KG, we
first convert the drug entities into semantically meaningful fixed-length vector
representations using KGE models and then feed these embeddings into a ML
classifier that can distinguish between interacting and non-interacting drug
pairs. In this research, we explore and compare different translation,
factorisation, and NN-based KGE algorithms to discover which model is the most
suitable to address the DDI link prediction problem.

\subsubsection{Translation and Factorisation-based Approach}

Recent research has released open-source libraries to facilitate the adaptation
of these KGE models in various applications. For example, Celebi et al.
\cite{celebi} used the OpenKE library \cite{openke} to train the TransE and
TransD models, whilst Karim et al. \cite{karim} used the Pytorch BigGraph
library \cite{pytorch} amongst other libraries. Moreover, in 2020, Amazon
launched DGL-KE \cite{dglke}, a high-performance, easy-to-use, and scalable
toolkit to generate some of the most popular KGEs from large graphs. This
library builds upon Deep Graph Learning (DGL) \cite{dgl}, an open-source library
to implement GNNs, and allows one to train a KGE model with a simple input
argument in the command line script. In addition, DGL-KE contains various
optimisations that accelerate KGE training on KGs. We used the DGL-KE library to
train and compare the TransE, TransR, RESCAL, DistMult and ComplEx KGE models.

Given the dense representations created by one of the translation or
factorisation based KGE methods, we concatenated the embedding vectors of each
drug in the pair and then fed them into a ML model. We used concatenation as a
feature aggregation operator since the concatenate operator performs better than
other operators, such as inner product and summation \cite{feng, kgnn}.

We adopted three different supervised ML algorithms with varying levels of
complexities; RF, Multi-Layer Perceptron (MLP) and LSTM. The RF and MLP are
available from scikit-learn \cite{scikit-learn}, whilst we built the LSTM model
using Keras \cite{keras}.

The parameters used for building our RF classifier were inspired by Celebi et
al. \cite{celebi}. The number of estimators is 200, and the maximum depth of the
tree is 8. On the other hand, we train the MLP classifiers using a single layer
of 100 neurons for a maximum of 100 epochs using the Adam \cite{adam} version of
stochastic gradient descent optimiser with a learning rate of 0.001. Finally, we
used Rectified Linear Unit (ReLU) as the activation function widely adopted in
deep learning research.

Our LSTM network contains two hidden layers of LSTM, one with 100 neurons and
another with 64 neurons. The output of each LSTM layer is then passed to the
dropout layer to regularise learning to avoid overfitting. The final output
layer has one neuron for forecasting whether an interaction exists. Every LSTM
layer expects a three-dimensional input, $s \times f \times t$, where $s$ is the
sample size, $t$ is the time-step, and $f$ is the feature size. In our case, $s$
is the number of interacting and non-interacting drug pairs, $t$ is one since we
are only using one feature, which is the KGE at one point in time, and $f$ is
the length of our concatenated feature vector of \textit{Drug A} and
\textit{Drug B}. The efficient Adam optimiser \cite{adam} is also applied with
the binary cross-entropy loss function, and the model is trained for 100 epochs.

\subsubsection{Graph Neural Network-based Approach}

\begin{figure}[h]
\centering
    \includegraphics[width=0.7\linewidth]{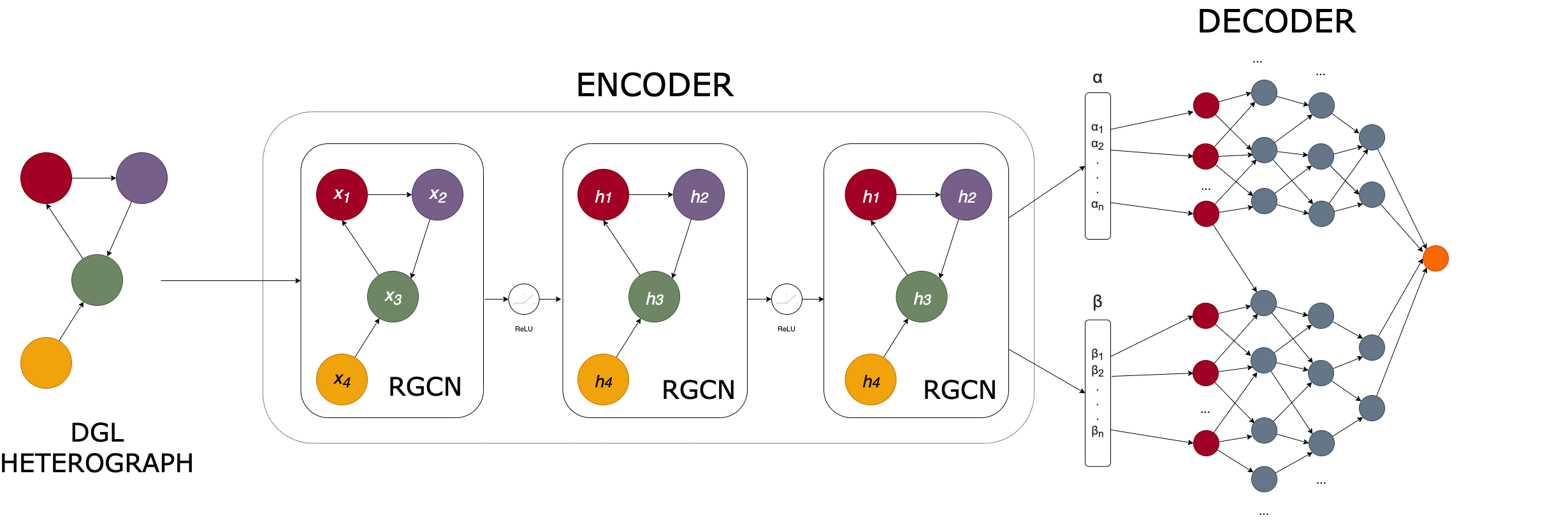}
    \caption{Graph Auto-Encoder Architecture}
    \label{fig:encoder}
\end{figure}

Additionally, we also implement a graph auto-encoder, illustrated in Figure
\ref{fig:encoder}, using PyTorch \cite{pytorch_library} and DGL \cite{dgl},
based on the approach presented by Schlichtkrull et al. \cite{zitnik}.

The encoder takes the positive training relations graph and additional node
feature vectors to the first R-GCN layer. Then, in each layer, the R-GCN
propagates the latent node feature information across the edges of the graph
while considering the relation type of the edge. A single layer of this model is
defined as;
\begin{equation}
    h_{i}^{(k+1)} = \phi(\sum_{r}\sum_{j \in N_{r}^{i}}c_{r}^{ij}W_{r}^{(k)}h_{j}^{(k)} + c_{r}^{i}h_{i}^{(k)})
\end{equation}
where $h_{i}^{(k)}$ is the hidden state of node $v_{i}$ in the $k^{th}$ layer of
the NN, $r$ is the relation type and matrix $W_{r}^{k}$ is a relation-type
weight matrix. $\phi$ denotes the activation function, which transforms the
representations to be used in the layer of the neural model, $c_{r}^{ij}$ and
$c_{r}^{i}$ are normalisation constants and $N_{r}^{i}$ denotes the set of
neighbours of node $v_{i}$ under relation $r$.

We stack a number of R-GCN layers using the \textit{HeteroGraphConv} module in
DGL. The resulting representation of the first R-GCN layer is accumulated and
passed through the ReLU activation function to produce the hidden state of each
node, which is then used as input to the second R-GCN layer.

On the other hand, the decoder aggregates our drug pairs’ entity
representations, generated by the R-GCN, into a single vector and computes a
probability for every potential edge in the graph that has a destination and
source node of type drug, using a simple NN.

In each iteration, we trained the encoder and decoder concurrently, and then
given a set of drug pairs and the ground-truth interaction values in the
training dataset, we applied the binary cross-entropy as the loss function to
assign higher probabilities to observed edges than to random non-edges. Finally,
to optimise all trainable parameters in the model, we used the Adam \cite{adam}
algorithm with a learning rate of 0.01. Gandhi et al. \cite{gandhi2021p3} noted
that for training tasks to achieve reasonable accuracy in GNNs, several 100s or
even 1000s of epochs are needed. For this reason, we set the minimum number of
epochs to 500.

\section{Evaluation}
This section presents the evaluation methodology we adopted to evaluate the
solutions to the objectives outlined in Section \ref{aim_objectives}.

To benchmark the performance of our classifier, we adopt a combination of the
evaluation metrics used by Ryu et al. \cite{ryu}, Celebi et al. \cite{celebi}
and Karim et al. \cite{karim}. These metrics include accuracy, precision,
recall, $F_{1}$ measure, Area Under the Receiver Operating Characteristic (AUC),
Area Under the Precision-Recall Curve (AUPR) and Matthias Correlation
Coefficient (MCC).

Like several other studies \cite{celebi, karim}, we decided to evaluate our
models on the known DDIs in the DrugBank version 5.1.8 dataset.

We used 70\% of the positive DDIs, in the DrugBank dataset, for the training,
30\% for evaluating the models, and 10\% from the training set was randomly used
for the validation. Since, in reality, we do not know all the DDIs, some of the
unknown DDIs may indeed be positive. However, regarding all unknown interactions
as the negative set, on the other hand, produces a data balance issue, affecting
performance metrics such as AUPR and $F_{1}$-score \cite{celebi}. Therefore,
similar to Celebi et al. \cite{celebi}, we randomly sampled an equivalent number
of negative samples from the unknown drug pairs as positive samples for each set.

\subsection{Knowledge Graph}

\begin{figure}[h!]
\centering
    \includegraphics[width=0.8\linewidth]{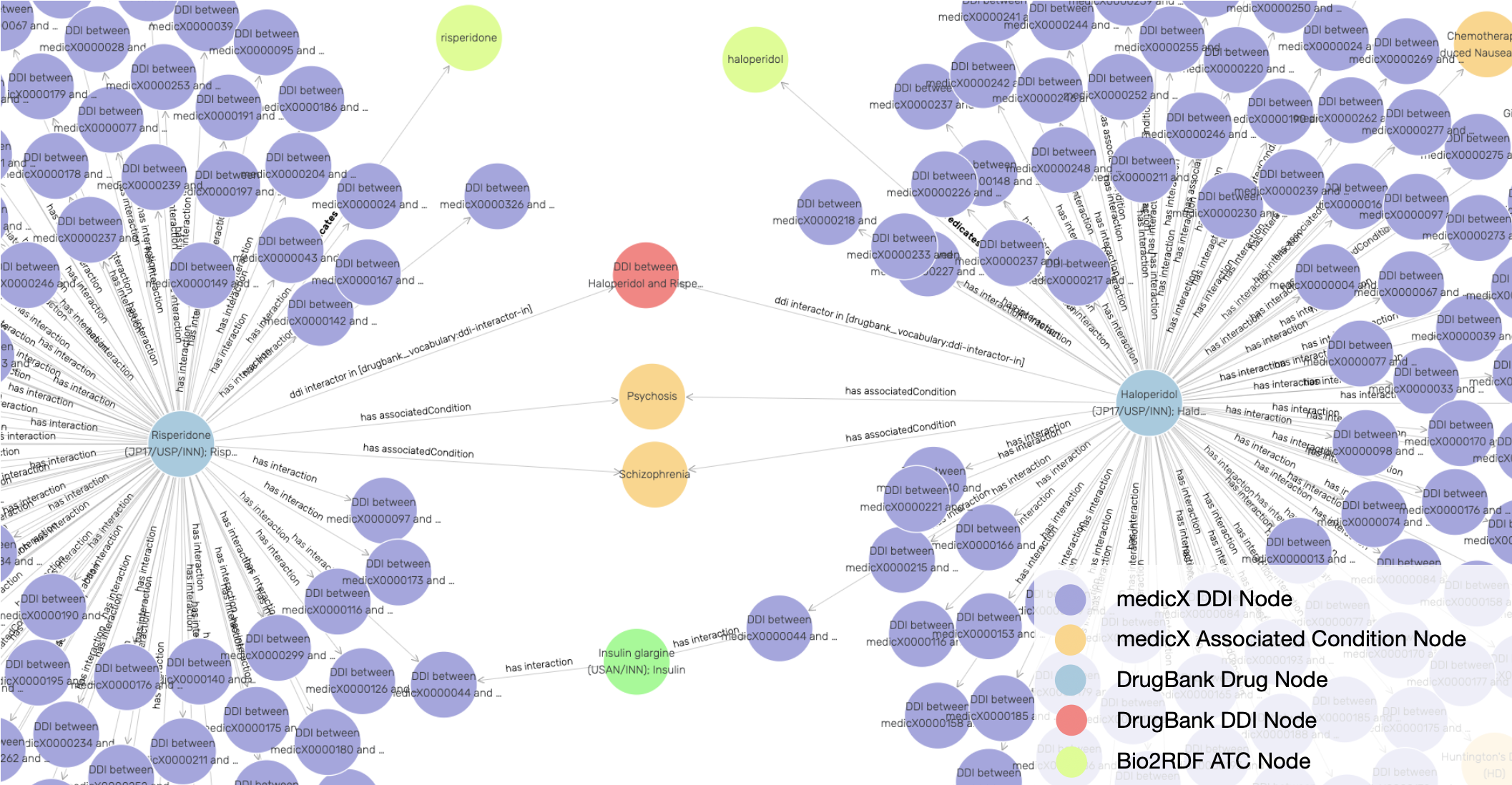}
    \caption{\textit{haloperidol} and \textit{risperidone} Sub-graph}
    \label{fig:kg_example}
\end{figure}

We built a KG comprising 18,322,022 nodes and 180,800,521 triples. Figure
\ref{fig:kg_example} depicts part of the KG that shows some of the nodes
connected to the \textit{haloperidol} and \textit{risperidone}, two
anti-psychotic medications, drug nodes. We note that the drugs interact since
they are linked to a common DDI node. In addition, we observed that both drugs
interact with \textit{insulin glargine}, a drug used to manage type I and type
II diabetes, and have two common associated conditions; schizophrenia and
psychosis.

Whilst we were analysing the relevancy of the KG in the real world, we compiled
a list of CQs, listed in Table \ref{tab:cq}, which we also used for evaluation
purposes. After building the KG, we wrote the questions as SPARQL queries,
similar to Belleau et al. \cite{bio2rdf}, and executed them in GraphDB using the
GraphDB Workbench. In addition to what we noticed earlier in Figure
\ref{fig:kg_example}, we discovered that our KG contains 11,107 and 7,402 DDIs
for \textit{haloperidol} and \textit{risperidone}, respectively. Also, we
uncovered that \textit{Baricitinib}, \textit{Bamlanivimab}, \textit{Moderna
COVID-19 Vaccine}, \textit{Pfizer-BioNTech Covid-19 Vaccine} and
\textit{Remdesivir} are all used to either treat or prevent Covid-19. Moreover,
the SPARQL endpoint listed 155 adverse reactions for \textit{haloperidol}. These
range from mild reactions such as increased sweating and nausea to serious ones,
including Parkinsonian Disorders and sudden death. When we executed the final
query, we observed that 36 drugs in our KG are anti-psychotics, which included
\textit{haloperidol} and \textit{risperidone}.

\begin{table}[h!]
\centering
\caption{Competency Questions}
\label{tab:cq}
\resizebox{0.6\linewidth}{!}{%
\begin{tabular}{ll}
\hline
CQ1 & Do \textit{haloperidol} and \textit{risperidone} interact with one another? \\
CQ2 & Which drugs interact with \textit{haloperidol} and \textit{risperidone}?    \\
CQ3 & Which drugs can be administered to treat Covid-19?                \\
CQ4 & What are side effects of \textit{haloperidol}?                     \\
CQ5 & Which drugs are antipsychotics?                           \\
 \hline
\end{tabular}%
}
\end{table}

As a result, these answers depict the similarities between \textit{haloperidol}
and \textit{risperidone}, and how these may lead to a greater additive or
synergistic effect (a DDI). These insights show the usefulness of KGs as they
can help us better understand the dynamics between entities, such as drugs in
this case, and why they may cause DDIs when consumed together. Moreover, with
the power of SPARQL together with the flexibility of RDF, we managed to
integrate previously disjoint systems. For example, using SPARQL we can navigate
from a DrugBank node to several ADRs within the SIDER dataset, which shows how
KGs can provide a unified view of varied unconnected data sources. Furthermore,
we observed that by having the essential relationships between concepts built
into our ontology, ontology languages, such as OWL and RDFS, enable automated
reasoning about data using a built-in reasoner in the graph database, improving
reusability and interoperability within a KG.

\subsection{Predicting Unknown Drug-Drug Interactions}

Among the different translation and factorisation-based KGE models, we noted
that the TransE and TransR models overall obtained the worst performance, whilst
DistMult and ComplEx obtained the best performance. This observation follows
since TransE cannot model complex and diverse relations, such as symmetric,
transitive, one-to-many, many-to-one, and many-to-many relations in KGs, and
TransR does not consider the diversity of node types.  Similar to Karim et al.
\cite{karim}, we noted that more complex models, such as the LSTM classifier,
outperformed the RF and MLP classifiers, in the best case resulting in an
$F_{1}$-score of 89.63\%, whilst the RF performed the worst, with the lowest
$F_{1}$-score of 72.23\%.

In addition, throughout our research, we observed that embedding quality
increases with higher dimensionality, and as a result, we set the embedding size
to 900. Moreover, we investigated the influence of the number of negatives per
positive training sample, known as negative sampling. We decreased the negative
sample size from 256 to 20, as Karim et al. \cite{karim} concluded that this
value obtained their best result and found that a larger sample size performs
better in our case. Ultimately, we generated a ComplEx-LSTM model that obtained
an $F_{1}$-score of 95.19\%.

We built several graph auto-encoders with varying numbers of R-GCN layers in the
encoder and hidden layers in the decoder. We observed that building a slightly
more complex architecture improves the performance of graph auto-encoders. For
example, by stacking three R-GCN layers in the encoder, and creating a DNN with
three hidden layers, instead of one in both components, the $F_{1}$-score
increased by 17.5\%. Moreover, by increasing the input and output embedding and
the hidden layer size in the R-GCN layer the model's performance improved by
almost 2\%.

Several researchers, such as Wang et al. \cite{wang2021survey}, noted that
instead of using randomly distributed initial node features, such as the
distribution proposed by Xavier et al. \cite{xavier}, one should use trained
embeddings as input to the first GNN layer. Therefore, we used the ComplEx KGE
model in DGL-KE to create embeddings of size 50 and used these features as
initial node features for the first R-GCN layer. This optimisation further
improved the $F_{1}$ score by at least 3.18\%, which confers that initial node
features also impact the quality of the embeddings that the R-GCN model
produces. The resulting graph auto-encoder achieved an $F_{1}$-score of 91.94\%.

Compared to the ComplEx KGE model, R-GCNs are memory intensive since they
require memory to store input data, weight parameters, and activations as input
propagates through the network \cite{zhou2021optimizing}. Additionally, they are
computationally more expensive since they introduce numerous parameters to the
convolutions taking place. Hence due to memory restrictions, we kept the
embedding size of the nodes relatively small compared to the embeddings produced
by the ComplEx KGE model. Although the ComplEx-LSTM model achieved an
$F_{1}$-score that is 3.25\% better than the graph auto-encoder, we conclude
that using a feature learning and extraction model such as R-GCN can generate
better embeddings compared to translation and factorisation-based models,
especially when using larger embeddings. This achievement is because of the GCNs
ability to consider all types of relations among the entity links, path, and
substructure information, using the incorporated CNNs.

\subsection{Performance Comparison}

We replicated and compared our ComplEx-LSTM and GNN-based models with four other
research works \cite{vilar, ryu, celebi, karim} mentioned in Section 2.2. Table
\ref{tab:compare_baselines} illustrates the respective results. We observed that
our two-stage approach (ComplEx-LSTM) and graph auto-encoder achieved better
results than all the other methods.

Our results showed that although Vilar et al.’s method does not obtain
competitive results due to its naive approach, representing drugs using their
respective SMILES notation is a good way, as can be seen from the results
achieved by the DeepDDI model. Nonetheless, as we previously mentioned, the
SMILES notation is not always available. Thus, such models cannot discover
potential DDIs concerning drugs without a SMILES notation or a SMILES that a
library, such as RDKit\footnote{\url{https://github.com/rdkit/rdkit}}, cannot
decode.

Furthermore, we observed a significant difference between the approach by Celebi
et al. \cite{celebi} and our ComplEx-LSTM model, which most likely occurred
because the authors used the RDF2Vec model and an RF. On the other hand,
although Karim et al. \cite{karim} also used the ComplEx KGE model and a DNN as
a predictor, our model achieved an increase of 12.04\%. This distinction is
because they used an embedding size of 300; they omitted DDIs while training the
embedding model and because of their Convolutional-LSTM model.

\begin{table}[]
\caption{Comparison with other models}
\label{tab:compare_baselines}
\resizebox{\linewidth}{!}{%
\begin{tabular}{lccccccc}
\hline
               & Precision & Recall & $F_{1}$-score & AUC    & AUPR   & MCC    & Accuracy \\ \hline
Vilar et al. \cite{vilar}          & 0.5138    & 0.5138 & 0.5132   & 0.5192 & 0.5156 & 0.0276 & 0.5138   \\
Ryu et al. \cite{ryu}            & 0.8958      & 0.8958   & 0.8958     & 0.9623 & 0.9621      & 0.7916      & 0.8958     \\
Celebi et al. \cite{celebi}         & 0.7887    & 0.782  & 0.7807   & 0.8627 & 0.8481 & 0.5707 & 0.7820   \\
Karim et al. \cite{karim}          & 0.8374    & 0.8321 & 0.8315   & 0.9178 & 0.9138 & 0.6695  & 0.8321   \\
Our ComplEx + LSTM & \textbf{0.9519}    & \textbf{0.9519} & \textbf{0.9519}   & \textbf{0.9897}  & \textbf{0.9901} & \textbf{0.9038} & \textbf{0.9519}   \\
Our Graph auto-encoder            & \textbf{0.9195}          & \textbf{0.9194}       & \textbf{0.9194}         & \textbf{0.9789}       & \textbf{0.9794}        & \textbf{0.8389}       & \textbf{0.9194}         \\ \hline
\end{tabular}
}
\end{table}

\begin{table}[]
\caption{Comparison with other models}
\label{tab:compare_baselines}
\resizebox{\linewidth}{!}{%
\begin{tabular}{lccccccc}
\hline
               & Precision & Recall & $F_{1}$-score & AUC    & AUPR   & MCC    & Accuracy \\ \hline
Vilar et al. (2012)          & 0.5138    & 0.5138 & 0.5132   & 0.5192 & 0.5156 & 0.0276 & 0.5138   \\
Ryu et al. (2018)            & 0.8958      & 0.8958   & 0.8958     & 0.9623 & 0.9621      & 0.7916      & 0.8958     \\
Celebi et al. (2018)        & 0.7887    & 0.782  & 0.7807   & 0.8627 & 0.8481 & 0.5707 & 0.7820   \\
Karim et al. (2019)          & 0.8374    & 0.8321 & 0.8315   & 0.9178 & 0.9138 & 0.6695  & 0.8321   \\
Our ComplEx + LSTM & 0.9519    & 0.9519 & 0.9519   & 0.9897  & 0.9901 & 0.9038 & 0.9519   \\
Our Graph auto-encoder            & \textbf{0.9195}          & \textbf{0.9194}       & \textbf{0.9194}         & \textbf{0.9789}       & \textbf{0.9794}        & \textbf{0.8389}       & \textbf{0.9194}         \\ \hline
\end{tabular}
}
\end{table}

We also evaluated our DDI prediction models for cold start drugs, with no known
DDI information in the training set to determine whether our models are adequate
for drug discovery applications, as was performed by Celebi et al.
\cite{celebi}. Our ComplEx-LSTM model obtains a recall of 89.29\% and a
precision of 54.59\% for interacting drug pairs when only one drug in the drug
pair is a new drug. On the other hand, when both drugs are cold start drugs, our
model obtained a precision of 50.52\% and a recall of 95.09\%. We noted that the
false-negative rate is low (high recall), and this is beneficial in most medical
practices since such a model can help professionals determine potentially
dangerous DDIs. Although the precision scores seem to be low, we consider the
results to be realistic in predicting the interactions for drugs with
insufficient interaction information. Hence, we can presume that our model can
be helpful in drug discovery applications.

Lastly, by altering the LSTM to handle more than two class labels and using the
trained ComplEx embeddings that obtained the best results in our binary
classification, we implemented a multi-classifier that, besides distinguishing
between positive and negative DDIs, can classify a positive DDI as one out of 86
positive DDI data types mentioned in the DeepDDI dataset. In Table
\ref{tab:multi_class}, we compare our results with those reported by Ryu et al.
\cite{ryu} and Celebi et al. \cite{celebi} in their paper. As a result, we
observed that, apart from a slight improvement and faster training, our model,
unlike DeepDDI, can predict DDIs of drugs that do not have the SMILES feature.

\begin{table}[h!]
\centering
\caption{Multi-class Classifiers comparison}
\label{tab:multi_class}
\resizebox{0.7\linewidth}{!}{%
\begin{tabular}{cccccc}
\hline
 & Precision & Recall & $F_{1}$-score & MCC & Accuracy \\ \hline
Ryu et al. \cite{ryu} & 0.9 & 0.85 & 0.86 & - & 0.92 \\ 
Celebi et al. \cite{celebi} & 0.82 & 0.87 & 0.85 & - & 0.84 \\ \hline
ComplEx - LSTM & \textbf{0.9086} & \textbf{0.8441} & \textbf{0.8649} & \textbf{0.9367} & \textbf{0.9549} \\\hline
\end{tabular}%
}
\end{table}

\section{Conclusion}

We managed to address both objectives we laid out in Section
\ref{aim_objectives}. We built a KG that can easily be queried to extract
knowledge regarding drugs, DDIs, side effects and indications and showed that
apart from acting as a data structure, KGs can be used by factorisation-based
KGE models and GNNs to generate high-quality feature representations that can be
used by ML algorithms or deeper networks for DDI prediction.

Throughout our evaluation, we learned that KGs can capture relationships between
drugs and graph databases are able to discern valuable insights on why drugs
might interact with one another based on links to common nodes such as side
effects and indications. Furthermore, KGE models provide better drug
representation than traditional feature extraction and selection methods, to
predict potential DDIs, as they do not rely on a limited number of drug
properties.

There is still work that can be done to improve the performance and efficiency
of our system. We plan to enrich the KG with other structured and unstructured
data from other medical sources, including PubMed, which is a search engine that
provides accessibility to biomedical literature. Moreover, we want to consult
professionals in the biomedical area to find out what other drug information is
helpful to include in our KG. For the second objective, we would like to extend
our graph auto-encoder to predict multiple DDI types by replacing the
interaction relations with the interaction types and defining a decoder that can
predict among them \cite{zitnik}. Finally, we noted that the negative set in our
evaluation consists of drug pairs, for which evidence regarding their
interaction has not been found yet, and may still interact with one another.
Therefore, we may have discrepancies within our dataset and cannot correctly
identify true negative and false positive drug pairs \cite{rohani}. Hence, we
want to train our models using a dataset containing only true negatives in their
negative set to create a more precise and reliable DDI predictor.




\bibliographystyle{unsrtnat}
\bibliography{reference}


\end{document}